# Loopy Belief Propagation for Approximate Inference: An Empirical Study

Kevin P. Murphy and Yair Weiss and Michael I. Jordan
Computer Science Division
University of California
Berkeley, CA 94705
{murphyk,yweiss,jordan}@cs.berkeley.edu

## Abstract

Recently, researchers have demonstrated that "loopy belief propagation" — the use of Pearl's polytree algorithm in a Bayesian network with loops — can perform well in the context of error-correcting codes. The most dramatic instance of this is the near Shannon-limit performance of "Turbo Codes" — codes whose decoding algorithm is equivalent to loopy belief propagation in a chain-structured Bayesian network.

In this paper we ask: is there something special about the error-correcting code context, or does loopy propagation work as an approximate inference scheme in a more general setting? We compare the marginals computed using loopy propagation to the exact ones in four Bayesian network architectures, including two real-world networks: ALARM and QMR. We find that the loopy beliefs often converge and when they do, they give a good approximation to the correct marginals. However, on the QMR network, the loopy beliefs oscillated and had no obvious relationship to the correct posteriors. We present some initial investigations into the cause of these oscillations, and show that some simple methods of preventing them lead to the wrong results.

## 1 Introduction

The task of calculating posterior marginals on nodes in an arbitrary Bayesian network is known to be NP-hard [5]. This is true even for the seemingly easier task of calculating approximate posteriors [6]. Nevertheless, due to the obvious practical importance of this task, there has been considerable interest in assessing the quality of different approximation schemes, in an attempt to delimit the types of networks and parameter regimes for which each scheme works best.

In this paper we investigate the approximation performance of "loopy belief propagation". This refers to using the well-known Pearl polytree algorithm [12] on a Bayesian network with loops (undirected cycles). The algorithm is an exact inference algorithm for singly-connected networks — the beliefs converge to the correct marginals in a number of iterations equal to the diameter of the graph.[1] However, as Pearl noted, the same algorithm will not give the correct beliefs for multiply connected networks:

> When loops are present, the network is no longer singly connected and local propagation schemes will invariably run into trouble ... If we ignore the existence of loops and permit the nodes to continue communicating with each other as if the network were singly connected, messages may circulate indefinitely around the loops and the process may not converge to a stable equilibrium ... Such oscillations do not normally occur in probabilistic networks ... which tend to bring all messages to some stable equilibrium as time goes on. However, this asymptotic equilibrium is not coherent, in the sense that it does not represent the posterior probabilities of all nodes of the network [12, p.195]

Despite these reservations, Pearl advocated the use of belief propagation in loopy networks as an approximation scheme (J. Pearl, personal communication) and exercise 4.7 in [12] investigates the quality of the approximation when it is applied to a particular loopy belief network.

Several groups have recently reported excellent experimental results by using this approximation scheme — by running algorithms equivalent to Pearl's algorithm on networks with loops. Perhaps the most dramatic instance of this performance is in an error correcting code scheme known as "Turbo Codes" [4]. These codes have been described as "the most exciting and potentially important development in coding theory in many years" [11] and have recently been shown [9, 10] to utilize an algorithm equivalent to belief propagation in a network with loops. Although there is widespread agreement in the coding community that these codes "represent a genuine, and perhaps historic, breakthrough" [11], a theoretical understanding of their performance has yet to be achieved. Yet McEliece et. al conjectured that the performance of loopy belief propagation on the Turbo code structure was a special case of a more general phenomenon:

> We believe there are general undiscovered theorems about the performance of belief propagation on loopy DAGs. These theorems, which may have nothing directly to do with coding or decoding will show that in some sense belief propagation "converges with high probability to a near-optimum value" of the desired belief on a class of loopy DAGs [10].

---

[1] This assumes parallel updating of all nodes. The algorithm can also be implemented in a centralized fashion in which case it converges in two iterations [13].

Progress in the analysis of loopy belief propagation has been made for the case of networks with a single loop [18, 19, 2, 1]. For the sum-product (or "belief update") version it can be shown that:

- Unless all the conditional probabilities are deterministic, belief propagation will converge.

- There is an analytic expression relating the correct marginals to the loopy marginals. The approximation error is related to the convergence rate of the messages — the faster the convergence the more exact the approximation.

- If the hidden nodes are binary, then thresholding the loopy beliefs is guaranteed to give the most probable assignment, even though the numerical value of the beliefs may be incorrect. This result only holds for nodes in the loop.

In the max-product (or "belief revision") version, Weiss [19] showed that (1) belief propagation may converge to a stable value or oscillate in a limit cycle and (2) if it converges then it is guaranteed to give the correct assignment of values to the hidden nodes. This result is independent of the arity of the nodes and whether the nodes are inside or outside the loop.

For the case of networks with multiple loops, Richardson [14] has analyzed the special case of Turbo codes. He has shown that fixed points of the sum-product version always exist, and has given sufficient conditions under which they will be unique and stable (although verifying these conditions may be difficult for large networks).

To summarize, what is currently known about loopy propagation is that (1) it works very well in an error-correcting code setting and (2) there are conditions for a single-loop network for which it can be guaranteed to work well. In this paper we investigate loopy propagation empirically under a wider range of conditions.

Is there something special about the error-correcting code setting, or does loopy propagation work as an approximation scheme for a wider range of networks?

## 2 The algorithm

For completeness, we briefly summarize Pearl's belief propagation algorithm. Each node $X$ computes a belief $BEL(x) = P(X = x|E)$, where $E$ denotes the observed evidence, by combining messages from its children $\lambda_{Y_j}(x)$ and messages from its parents $\pi_X(u_k)$. (Following Peot and Shachter [13], we incorporate evidence by letting a node send a message to itself, $\lambda_X(x)$.)

$$BEL(x) = \alpha\lambda(x)\pi(x) \quad (1)$$

where:

$$\lambda^{(t)}(x) = \lambda_X(x) \prod_j \lambda_{Y_j}^{(t)}(x) \quad (2)$$

and:

$$\pi^{(t)}(x) = \sum_u P(X = x|U = u) \prod_k \pi_X^{(t)}(u_k) \quad (3)$$

The message $X$ passes to its parent $U_i$ is given by:

$$\lambda_X^{(t+1)}(u_i) = \alpha \sum_x \lambda^{(t)}(x) \sum_{u_k:k \neq i} P(x|u) \prod_{k \neq i} \pi_X^{(t)}(u_k) \quad (4)$$

and the message $X$ sends to its child $Y_j$ is given by:

$$\pi_{Y_j}^{(t+1)}(x) = \alpha\pi^{(t)}(x)\lambda_X(x) \prod_{k \neq j} \lambda_{Y_k}^{(t)}(x) \quad (5)$$

For noisy-or links between parents and children, there exists an analytic expression for $\pi(x)$ and $\lambda_X(u_i)$ that avoids the exhaustive enumeration over parent configurations [12].

We made a slight modification to the update rules in that we normalized both $\lambda$ and $\pi$ messages at each iteration. As Pearl [12] pointed out, normalizing the messages makes no difference to the final beliefs but avoids numerical underflow.

Nodes were updated in parallel: at each iteration all nodes calculated their outgoing messages based on the incoming messages of their neighbors from the previous iteration. The messages were said to converge if none of the beliefs in successive iterations changed by more than a small threshold ($10^{-4}$). All messages were initialized to a vector of ones; random initialization yielded similar results, since the initial conditions rapidly get "washed out".

For comparison, we also implemented likelihood weighting [17], which is a simple form of importance sampling. Like any sampling algorithm, the errors can be driven towards zero by running the algorithm for long enough; in this paper, we usually used 200 samples, so that the total amount of computation time was roughly comparable (to within an order of magnitude) to loopy propagation. We did not implement some of the more sophisticated versions of likelihood weighting, such as Markov blanket scoring [16], since our goal in this paper was to evaluate loopy propagation rather than exhaustively compare the performance of alternative algorithms. (For a more careful evaluation of likelihood weighted sampling in the case of the QMR network, see [8].)

## 3   The networks

We used two synthetic networks, PYRAMID and toyQMR, and two real world networks, ALARM and QMR. The synthetic networks are sufficiently small that we can perform exact inference, using the junction tree algorithm. This allows us to measure the accuracy of the approximation scheme. All the networks have many loops of different sizes.

### 3.1   The PYRAMID network

Figure 1 shows the structure of the PYRAMID network. This is a multilayered hierarchical network with local connections between each layer and observations only at the bottom layer. We chose this structure because networks of this type are often used in image analysis — the bottom layer would correspond to pixels (see for example [15]).

All nodes were binary and the conditional probabilities were represented by tables — entries in the conditional probability tables (CPTs) were chosen uniformly in the range $[0, 1]$.

### 3.2   The toyQMR network

Figure 2 shows the structure of a "toyQMR" network. This network is meant to represent the types of networks that arise in medical diagnosis — hidden diseases in the top layer and observed symptoms in the bottom layer. Here we randomized over structure and parameters — for each experiment the parents of each node in the bottom layer was a randomly chosen subset of the nodes in the top layer. The parents subset was chosen using a simple procedure — each parent-child link was either present or absent with a probability of 0.5.

All nodes were binary and the conditional probabilities of the leaves were represented by a noisy-or:

$$P(\text{Child} = 0|\text{Parents}) = e^{-\theta_0 - \sum_i \theta_i \text{Parent}_i} \quad (6)$$

where $\theta_0$ represents the "leak" term.

The links $\theta_i$ were chosen uniformly in the range $[0, 1]$ while $\theta_0$ was chosen uniformly in the range $[0, 0.01]$ (hence the leaks are inhibited with very high probability). The top layer had prior probabilities represented as CPTs and they were chosen uniformly in the range $[0, 1]$.

### 3.3   The ALARM network

Figure 3 shows the structure of the ALARM network — a Bayesian network for monitoring patients in intensive care. This network was used by [3] to compare various inference algorithms. The arity of the nodes ranges from two to four and all conditional distributions are represented by tables. The structure and the CPTs were downloaded from Nir Friedman's Bayesian network repository at: www.cs.huji.ac.il/~nir.

### 3.4   The QMR-DT network

The QMR-DT is a bipartite network whose structure is the same as that shown in figure 2 but the size is much larger. There are approximately 600 diseases and approximately 4000 findin nodes, with a number of observed findings that varies per case. Due to the form of the noisy-or CPTs the complexity of inference is exponential in the number of positive findings [7]. Following [8], we focused on the four CPC cases for which the number of positive findings is less than 20, so that exact inference is possible (using the QUICKSCORE algorithm [7]).

## 4   Results

### 4.1   Initial experiments

The experimental protocol for the PYRAMID network was as follows. For each experimental run, we first generated random CPTs. We then sampled from the joint distribution defined by the network and clamped the observed nodes (all nodes in the bottom layer) to their sampled value. Given a structure and observations, we then ran three inference algorithms — junction tree, loopy belief propagation and sampling.

We found that loopy belief propagation always converged in this case with the average number of iterations equal to 10.2. Figure 4(a) shows the correlation plot between the exact marginals (calculated using junction tree) and the loopy marginals ($BEL(x)$ in equation 1 at convergence). For comparison, figure 4(b) shows the correlation between likelihood weighting and the correct marginals. Note that the sampler has been run for 20 times as many iterations as loopy propagation.

The experimental protocol for the toyQMR network was similar to that of the PYRAMID network except that we randomized over structure as well. Again we found that loopy belief propagation always converged, with the average number of iterations equal to 8.65. Figure 5 shows the two correlation plots.

The protocol for the ALARM network experiments differed from the previous two in that the structure and parameters were fixed — only the observed evidence differed between experimental runs. We assumed that all leaf nodes were observed and calculated the pos-terior marginals of all other nodes. Again we found that loopy belief propagation always converged with the average number of iterations equal to 14.55. Figure 6 shows the correlation plots. With 200 samples, the correlation for likelihood weighting is rather weak, perhaps due to the larger arity of some of the nodes (and hence the larger state space); after 1000 samples, the correlation improves considerably.

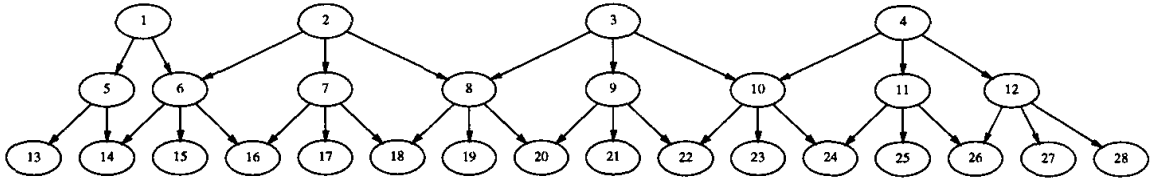

Figure 1: The structure of the PYRAMID network. All nodes are binary and observations appear only on the bottom layer. Such networks occur often in image analysis where the bottom layer would correspond to pixels.

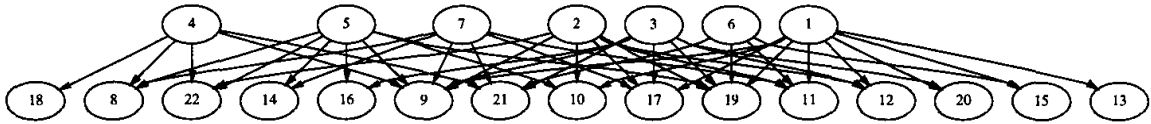

Figure 2: The structure of a toyQMR network. This is a bipartite structure where the conditional distributions of the leaves are noisy-or's. The network shown represents one sample from randomly generated structures where the parents of each symptom were a random subset of the diseases.

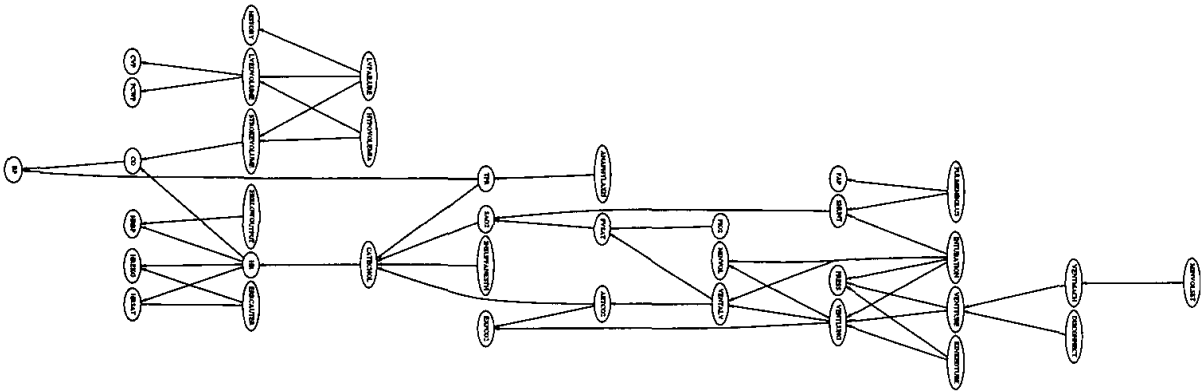

Figure 3: The structure of the ALARM network — a network constructed by medical experts for monitoring patients in intensive care.

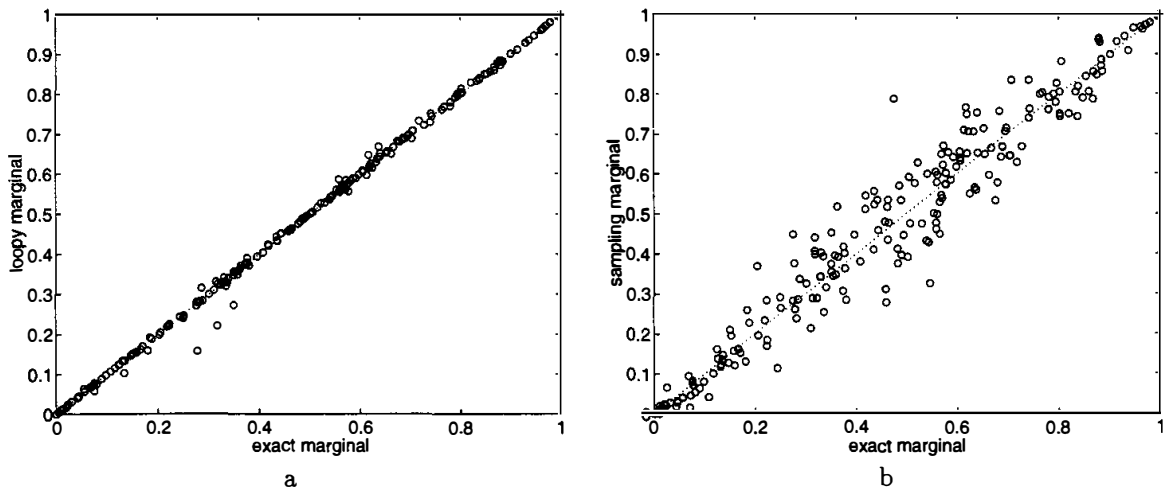

Figure 4: Correlation plots between the correct and approximate beliefs for the PYRAMID network, using (a) loopy propagation and (b) likelihood weighting with 200 samples.

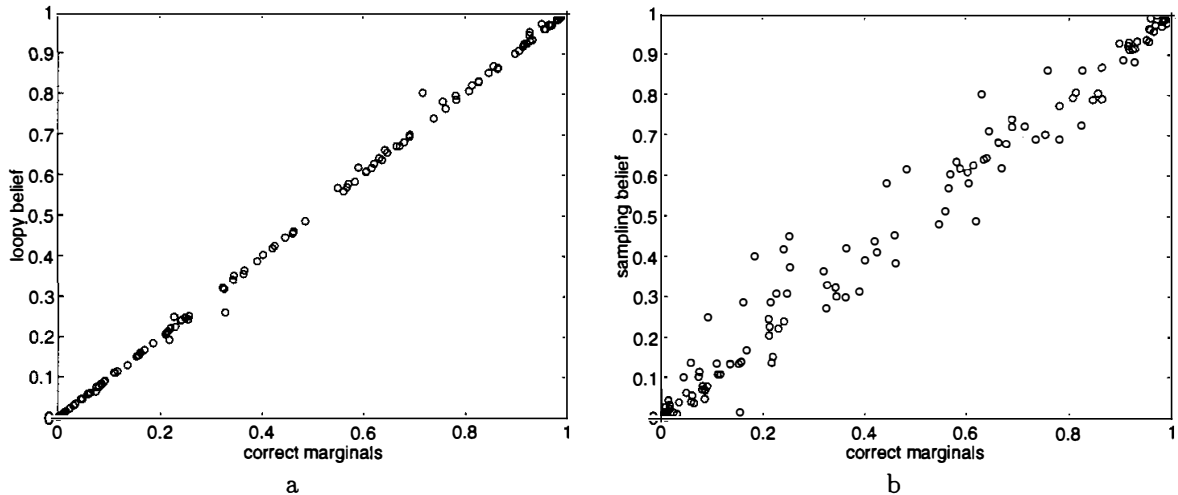

Figure 5: Correlation plots between the correct and approximate beliefs for the toyQMR network, using (a) loopy propagation and (b) likelihood weighting with 200 samples.

The results presented up until now show that loopy propagation performs well for a variety of architectures involving multiple loops. We now present results for the QMR-DT network which are not as favorable.

In the QMR-DT network there was no randomization. We used the fixed structure and calculated posteriors for the four cases for which posteriors have been calculated exactly by Heckerman [7]. For none of these four cases did loopy propagation converge. Rather, the loopy marginal oscillated between two quite distinct values for nearly all nodes. Figure 7(a) shows three such marginals. After two iterations the marginal seems to converge to a limit cycle with period two. In Figure 7(b) it seems that the correct posteriors always lie inside the interval defined by the limit cycle. However, this is not always the case (except, of course, when the interval is 0 to 1!).

### 4.2    What causes convergence versus oscillation?

What our initial experiments show is that loopy propagation does a good job of approximating the correct posteriors if it converges. Unfortunately, on the most challenging case — the QMR-DT network — the algorithm did not converge. We wanted to see if this oscillatory behavior in the QMR-DT case was related to the size of the network — does loopy propagation tend to converge less for large networks than small networks?

To investigate this question, we tried to cause oscillation in the toyQMR network. We first asked what, besides the size, is different between toyQMR and real QMR? An obvious difference is in the parameter values — while the CPTs for toyQMR are random, the real QMR parameters are not. In particular, the prior probability of a disease node being on is extremely low in the real QMR (typically of the order of $10^{-3}$).

Would low priors cause oscillations in the toyQMR case? To answer this question we repeated the experiments reported in the previous section but rather than having the prior probability of each node be randomly selected in the range $[0,1]$ we selected the prior uniformly in the range $[0,U]$ and varied $U$. Unlike the previous simulations we did not set the observed nodes by sampling from the joint — for low priors all the findings would be negative and inference would be trivial. Rather each finding was independently set to positive or negative. Figure 8 shows the results — for small priors the toyQMR network does not converge and we find the same oscillatory behavior as in the real QMR network case.

If indeed small priors are responsible for the oscillation, then we would expect the real QMR network to converge if the priors were sampled randomly in the range $[0,1]$. To check this, we reran loopy propagation on the full QMR network with the four tractable cases but changed the priors to be randomly sampled in the range $[0,1]$. All other parameters remained the same as in the real QMR network. Now we found convergence on all four cases and the beliefs gave a very good correlation with the ones calculated using QUICKSCORE.

Small priors are not the only thing that causes oscillation. Small weights can, too. The effect of both is to reduce the probability of positive findings. We conjectured that the reason for the oscillations is that the observed data, which has many positive findings, is very untypical in this parameter regime. This would also explain why we didn't find oscillations in the other examples, where the data was sampled from the joint distribution encoded by the network.

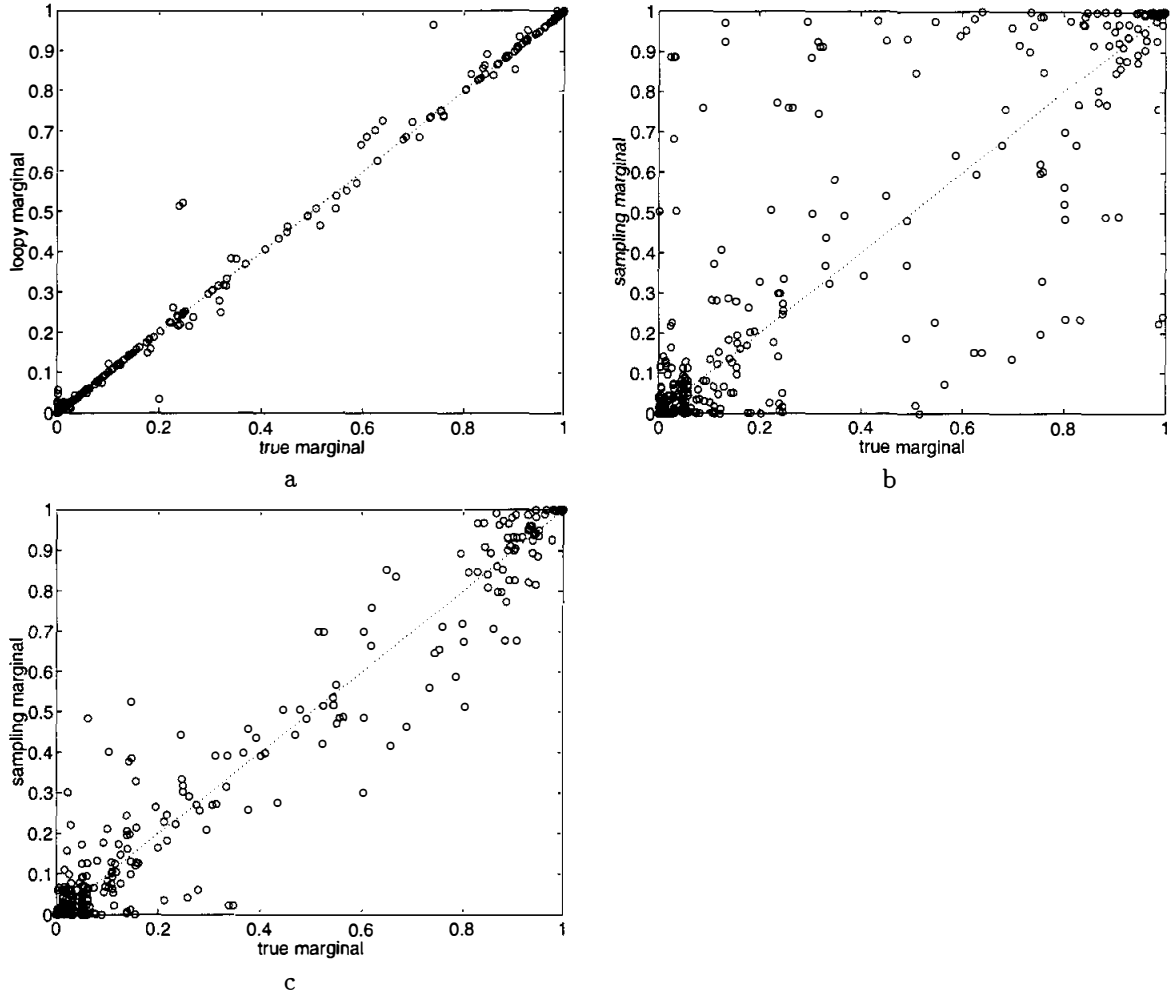

Figure 6: Correlation plots between the correct and approximate beliefs on the ALARM network, using (a) loopy propagation, (b) likelihood weighting with 200 samples, and (c) likelihood weighting with 1000 samples.

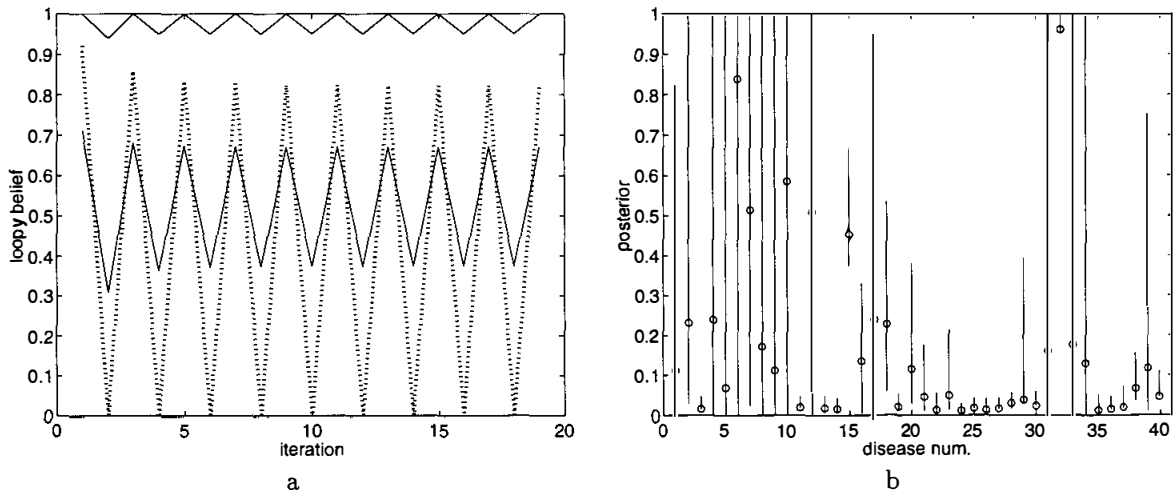

Figure 7: (a) The marginal posteriors for three of the nodes in the QMR-DT network. Note the limit cycle behavior. (b) The exact marginals are represented by the circles; the ends of the "error bars" represent the loopy marginals at the last two iterations. We only plot the diseases which had non-negligible posterior probability.

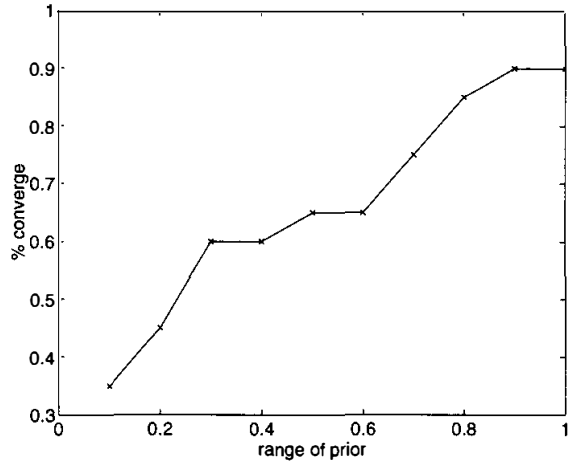

Figure 8: The probability of convergence in the toyQMR case as the upper bound on the priors of the diseases is increased. When the prior is small (similar to the real QMR regime) toyQMR converges quite rarely. This suggests that the failure of convergence in the real QMR cases is related to the low prior

To test this hypothesis, we reparameterized the pyramid network as follows: we set the prior probability of the "1" state of the root nodes to 0.9, and we utilized the noisy-OR model for the other nodes with a small (0.1) inhibition probability (apart from the leak term, which we inhibited with probability 0.9). This parameterization has the effect of propagating 1's from the top layer to the bottom. Thus the true marginal at each leaf is approximately $(0.1, 0.9)$, i.e., the leaf is 1 with high probability. We then generated untypical evidence at the leaves by sampling from the uniform distribution, $(0.5, 0.5)$, or from the skewed distribution $(0.9, 0.1)$. We found that loopy propagation still converged[2], and that, as before, the marginals to which it converged were highly correlated with the correct marginals. Thus there must be some other explanation, besides untypicality of the evidence, for the oscillations observed in QMR.

### 4.3 Can we fix oscillations easily?

When loopy propagation oscillates between two steady states it seems reasonable to try to find a way to combine the two values. The simplest thing to do is to average them. Unfortunately, this gave very poor results, since the correct posteriors do not usually lie in the midpoint of the interval (cf. Figure 7(b)).

---

[2]More precisely, we found that with a convergence threshold of $10^{-4}$, 98 out of 100 cases converged; when we lowered the threshold to $10^{-3}$, all 100 cases converged.

We also tried to avoid oscillations by using "momentum"; replacing the messages that were sent at time $t$ with a weighted average of the messages at times $t$ and $t - 1$. That is, we replaced the reference to $\lambda_{Y_j}^{(t)}$ in Equation 2 with

$$(1 - \mu)\lambda_{Y_j}(x)^{(t)} + \mu\lambda_{Y_j}(x)^{(t-1)} \qquad (7)$$

and similarly for $\pi_X^{(t)}$ in Equation 3, where $0 \leq \mu \leq 1$ is the momentum term. It is easy to show that if the modified system of equations converges to a fixed point $F$, then $F$ is also a fixed point of the original system (since if $\lambda_{Y_j}^{(t)} = \lambda_{Y_j}^{(t-1)}$, then Equation 7 yields $\lambda_{Y_j}^{(t)}$).

In the experiments for which loopy propagation converged (PYRAMID, toyQMR and ALARM), we found that adding the momentum term did not change the results — the beliefs that resulted were the same beliefs found without momentum. In the experiments which did not converge (toyQMR with small priors and real QMR), we found that momentum significantly reduced the chance of oscillation. However, in several cases the beliefs to which the algorithm converged were quite inaccurate — see Figure 9.

## 5 Discussion

The experimental results presented here suggest that loopy propagation can yield accurate posterior marginals in a more general setting than that of error-correcting coding — the PYRAMID, toyQMR and ALARM networks are quite different from the error-correcting coding graphs yet the loopy beliefs show high correlation with the correct marginals.

In error-correcting codes the posterior is typically highly peaked and one might think that this feature is necessary for the good performance of loopy propagation. Our results suggest that is not the case — in none of our simulations were the posteriors highly peaked around a single joint configuration. If the probability mass was concentrated at a single point the marginal probabilities should all be near zero or one; this is clearly not the case as can be seen in the figures.

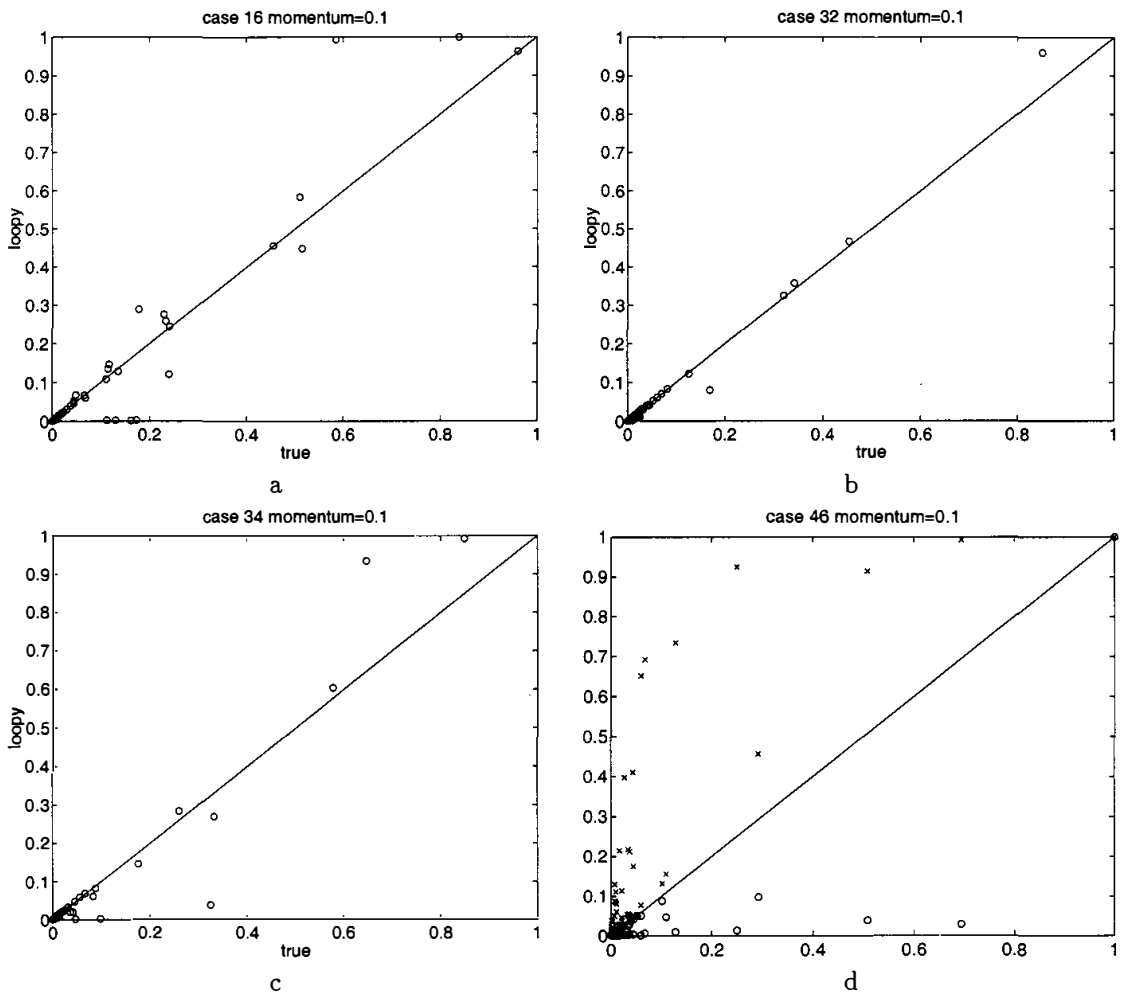

Figure 9: Correlation plots between the correct and approximate beliefs on the QMR-DT network and loopy propagation with momentum for the four tractable cases. With $\mu = 0.1$, we get convergence at the $10^{-3}$ level for cases 16, 32 and 34, but not for case 46. (a), (b) and (c) plots the results for the first three cases at convergence (usually 10–15 iterations): note several highly uncorrelated points. (d) plots the results for case 46 at time 20 ('o') and time 19 ('x'). This 'flip-flop' behavior around the diagonal is typical for the non-converging QMR cases.

It might be expected that loopy propagation would only work well for graphs with large loops. However, our results, and previous results on turbo codes, show that loopy propagation can also work well for graphs with many small loops.

At the same time, our experimental results suggest a cautionary note about loopy propagation, showing that the marginals may exhibit oscillations that have very little correlation with the correct marginals. We presented some preliminary results investigating the cause of the oscillations and showed that it is not simply a matter of the size of the network or the number of parents. Rather the same structure with different parameter values may oscillate or exhibit stable behavior.

For all our simulations, we found that when loopy propagation converges, it gives a surprisingly good approximation to the correct marginals. Since the distinction between convergence and oscillation is easy to make after a small number of iterations, this may suggest a way of checking whether loopy propagation is appropriate for a given problem.

**Acknowledgements**

We thank Tommi Jaakkola, David Heckerman and David MacKay for useful discussions. We also thank Randy Miller and the University of Pittsburgh for the use of the QMR-DT database. Supported by MURI-ARO DAAH04-96-1-0341.

## References

[1] J. M. Agosta. The structure of Bayes networks for visual recognition. In *UAI*, volume 4, pages 397–405, 1990.

[2] S.M. Aji, G.B. Horn, and R.J. McEliece. On the convergence of iterative decoding on graphs with a single cycle. In *Proc. 1998 ISIT*, 1998.

[3] I. Beinlich, G. Suermondt, R. Chavez, and G. Cooper. The alarm monitoring system: A case study with two probabilistic inference techniques for belief networks. In *Proc. 2'nd European Conf. on AI and Medicine*, 1989.

[4] C. Berrou, A. Glavieux, and P. Thitimajshima. Near Shannon limit error-correcting coding and decoding: Turbo codes. In *Proc. IEEE International Communications Conference '93*, 1993.

[5] G. Cooper. The computational complexity of probabilistic inference using Bayesian belief networks. *Artificial Intelligence*, 42:393–405, 1990.

[6] P. Dagum and M. Luby. Aproximate probabilistic inference in Bayesian networks in NP hard. *Artificial Intelligence*, 60:141–153, 1993.

[7] D. Heckerman. A tractable inference algorithm for diagnosing multiple diseases. In *Proc. Fifth Conf. on Uncertainty in AI*, 1989.

[8] T.S. Jaakkola and M.I. Jordan. Variational probabilistic inference and the QMR-DT network. *JAIR*, 10, 1999.

[9] F. R. Kschischang and B. J. Frey. Iterative decoding of compound codes by probability propagation in graphical models. *IEEE Journal on Selected Areas in Communication*, 16(2):219–230, 1998.

[10] R.J. McEliece, D.J.C. MacKay, and J.F. Cheng. Turbo decoding as as an instance of Pearl's 'belief propagation' algorithm. *IEEE Journal on Selected Areas in Communication*, 16(2):140–152, 1998.

[11] R.J. McEliece, E. Rodemich, and J.F. Cheng. The Turbo decision algorithm. In *Proc. 33rd Allerton Conference on Communications, Control and Computing*, pages 366–379, Monticello, IL, 1995.

[12] Judea Pearl. *Probabilistic Reasoning in Intelligent Systems: Networks of Plausible Inference*. Morgan Kaufmann, 1988.

[13] M.A. Peot and R.D. Shachter. Fusion and propagation with multiple observations in belief networks. *Artificial Intelligence*, 48:299–318, 1991.

[14] Thomas Richardson. The geometry of turbo-decoding dynamics. *IEEE Trans. on Info. Theory*, 1999. To appear.

[15] L.K. Saul, T. Jaakkola, and M.I. Jordan. Mean field theory for sigmoid belief networks. *JAIR*, 4:61–76, 1996.

[16] R. D. Shachter and M. A. Peot. Simulation approaches to general probabilistic inference on belief networks. In *Uncertainty in AI*, volume 5, 1990.

[17] M. Shwe and G. Cooper. An empirical analysis of likelihood-weighting simulation on a large, multiply connected medical belief network. *Computers and Biomedical Research*, 24:453–475, 1991.

[18] Y. Weiss. Belief propagation and revision in networks with loops. Technical Report 1616, MIT AI lab, 1997.

[19] Y. Weiss. Correctness of local probability propagation in graphical models with loops. *Neural Computation*, to appear, 1999.

# Learning Bayesian Networks from Incomplete Data with Stochastic Search Algorithms

**James W. Myers**
George Mason University
Fairfax, VA 22032-4444
myers29@erols.com

**Kathryn Blackmond Laskey**
George Mason University
Fairfax, VA 22032-4444
klaskey@gmu.edu

**Tod Levitt**
IET
Setauket, NY 11733
tlevitt@iet.com

**Abstract**

This paper describes stochastic search approaches, including a new stochastic algorithm and an adaptive mutation operator, for learning Bayesian networks from incomplete data. This problem is characterized by a huge solution space with a highly multimodal landscape. State-of-the-art approaches all involve using deterministic approaches such as the expectation-maximization algorithm. These approaches are guaranteed to find local maxima, but do not explore the landscape for other modes. Our approach evolves structure and the missing data. We compare our stochastic algorithms and show they all produce accurate results.

## 1 INTRODUCTION

Bayesian networks are growing in popularity as the model of choice of many AI researchers for problems involving reasoning under uncertainty. They have been implemented in applications in areas such as medical diagnostics, classification systems, software agents for personal assistants, multisensor fusion, and legal analysis of trials. Until recently, the standard approach to constructing belief networks was a labor-intensive process of eliciting knowledge from experts. Methods for capturing available data to construct Bayesian networks or to refine an expert-provided network promise to greatly improve both the efficiency of knowledge engineering and the accuracy of the models. For this reason, learning Bayesian networks from data has become an increasingly active area of research. Most of the research to date has relied on the assumption that data are complete; that is, the values of all variables are known for all cases in the database. This assumption is not very realistic since most real world situations involve incomplete information.

Learning a Bayesian network can be decomposed into the problem of learning the graph structure and learning the parameters. The first attempts at treating incomplete data involved learning the parameters of a fixed network structure [Lauritzen 1995]. Very recently, researchers have begun to tackle the problem of learning the structure of the network from incomplete data. A major stumbling block in this research is that when information is missing, closed form expressions do not exist for the scoring metric used to evaluate the network structures. This has led many researchers down the path of estimating the score using parametric approaches such as the expectation-maximization (EM) algorithm [Dempster, Laird et al. 1977], [Friedman 1998]. The EM algorithm is a proven approach for dealing with incomplete information when building statistical models [Little and Rubin 1987]. EM and related algorithms show promise. However, it has been noted [Friedman 1998] that the search space is large and multimodal, and deterministic search algorithms are prone to find local optima. Multiple restarts have been suggested as a way to deal with this problem.

An obvious choice to combat the problem of "getting stuck" on local maxima is to use a stochastic search method. This paper explores the use of evolutionary algorithms (EA) and Markov chain Monte Carlo (MCMC) algorithms for learning Bayesian networks from incomplete data. We also introduce an algorithm, the Evolutionary Markov Chain Monte Carlo (EMCMC) algorithm, which combines the advantages of the EA and MCMC, which we believe, advances the state of the art for both EA